\documentclass[runningheads]{llncs}
\usepackage{multirow}
\usepackage{booktabs}
\usepackage{makecell}
\usepackage{array}
\usepackage{arydshln}
\usepackage[T1]{fontenc}
\usepackage{pifont}
\usepackage{graphicx,verbatim}
\usepackage{amsmath,amssymb,amsfonts}
\begin{document}
\newsavebox\CBox
\def\textBF#1{\sbox\CBox{#1}\resizebox{\wd\CBox}{\ht\CBox}{\textbf{#1}}}
\title{TEDi: Temporal Memory-Enhanced and Denoising Transformer for Surgical Instrument Segmentation}
%

\author{
Jiahong Yuan\inst{1}\textsuperscript{*}
\and
Weiming Mi\inst{2}\textsuperscript{*}
\and
Tao Zhang\inst{1}\textsuperscript{**}
\and
Haoyin Zhou\inst{2}\textsuperscript{**}
}

\authorrunning{J. Yuan et al.}

\institute{
Department of Automation, Tsinghua University, Beijing, China\\
\email{yuanjh25@mails.tsinghua.edu.cn, taozhang@tsinghua.edu.cn}
\and
Surgical Planning Laboratory, Brigham and Women's Hospital, Harvard Medical School, United States\\
\email{\{wemi,zhouhaoyin\}@bwh.harvard.edu}
}

\maketitle

\begingroup
\renewcommand{\thefootnote}{*}
\footnotetext{Co-first authors}
\renewcommand{\thefootnote}{**}
\footnotetext{Corresponding authors}
\endgroup

\begin{abstract}
Query-based segmentation methods have shown promising potential for surgical instrument segmentation and recognition, which is essential for scene understanding and downstream tasks in computer-assisted surgery. However, most existing approaches predominantly rely on per-frame predictions and overlook cross-frame temporal priors as well as temporal-consistency constraints. This limitation often leads to unstable query representations and suboptimal category recognition. In this paper, we propose \textbf{TEDi}, a \textbf{T}emporal memory-\textbf{E}nhanced and
\textbf{D}enoising transformer for surgical \textbf{i}nstrument segmentation
that addresses these issues through Memory Search Enhancement and Temporal Consistency Denoising.
The former introduces a query-level memory bank and a memory search enhancement encoder to retrieve discriminative representations from historical frames, enriching current-frame features. The latter constructs a temporally consistent reference as a cross-frame semantic anchor to suppress temporally unstable predictions and promote semantic coherence across frames. Extensive experiments on two benchmark datasets, EndoVis 2017 and EndoVis 2018, demonstrate that TEDi consistently outperforms state-of-the-art methods, highlighting its potential to further advance computer-assisted surgery. The code will be released after acceptance.
\end{abstract}

\section{Introduction}
Robot-assisted minimally invasive surgery (RAMIS) leverages dexterous articulated instruments and high-fidelity endoscopic imaging to enable precise manipulation while reducing surgical trauma and accelerating postoperative recovery~\cite{dagnino2024robot,westebring2008haptics}. 
Accurate surgical instrument segmentation is fundamental to surgical scene understanding~\cite{huang2025surgtpgs,ayobi2025pixel}, supporting downstream tasks such as instrument tracking~\cite{sestini2021kinematic}, pose estimation~\cite{rai2025surgpose}, and trajectory prediction~\cite{toussaint2021co}, and serving as a critical component for next-generation surgical robotic systems~\cite{de2025scaling,nagy2019dvrk,lu2021toward}.

Early methods formulated instrument segmentation as pixel-wise classifica-
tion, which often suffers from spatial class inconsistency~\cite{shvets2018automatic,zhao2020learning}. 
Recently, query-based segmentation (QBS) frameworks, such as Mask2Former~\cite{cheng2022masked}, have gained increasing attention due to their unified set-prediction paradigm that jointly
models instance masks and categories. Several studies have adapted QBS models to surgical scenarios. ISINet~\cite{gonzalez2020isinet} builds upon Mask R-CNN~\cite{he2017mask} for instrument segmentation, while other approaches improve QBS-based models by enhancing the discriminability of query representations~\cite{baby2023forks} and improving query initialization~\cite{dhanakshirur2023learnable}. Beyond single-frame prediction, temporal cues have been introduced to further improve performance and cross-frame consistency, including leveraging historical queries for tracking~\cite{zhao2022trasetr} and extending Mask2Former with video Transformers~\cite{ayobi2023MATIS} or spatio-temporal context modeling to strengthen temporal consistency and robustness~\cite{wang2025LACOSTE}.

However, empirical evidence suggests that, in surgical videos, QBS models typically achieve accurate localization while remaining prone to severe category prediction errors~\cite{ayobi2023MATIS,wang2025LACOSTE}. These errors manifest as (i) misclassification among visually confusable instruments and (ii) inconsistent class predictions for the same instance across adjacent frames, resulting in temporally unstable semantics. This indicates that query representations learned primarily under single-frame supervision are insufficient to exploit temporal cues as stable priors for fine-grained category discrimination.

To address these limitations, we propose a temporal-enhanced framework, TEDi, for surgical instrument segmentation, built upon two key principles: \textbf{Memory Search Enhancement (MSE)} and \textbf{Temporal Consistency Denoising (TCD)}. 
By explicitly mining temporal cues, TEDi systematically reduces category prediction errors and improves temporal semantic stability.

Our main contributions are summarized as follows:
\begin{itemize}
    \item  We propose \textbf{query-level memory bank}  and \textbf{memory search enhancement encoder} that retrieve target representations from historical frames to provide cross-frame semantic priors for current-frame segmentation.

    \item We develop a \textbf{window-based temporal-consistency denoising decoder} that constructs a temporally stable reference and enforces identity-aligned temporal correction to suppress unreliable query representations.
\end{itemize}

Extensive experiments on two EndoVis benchmark datasets demonstrate that our method achieves competitive or superior performance compared with state-of-the-art approaches, with particularly notable improvements on confusable categories and temporal stability.

\section{Method}
Our framework follows a query-centric pipeline. For each frame, we first extract instance-aware query embeddings and refine them via Memory Search Enhancement (MSE), which retrieves cross-frame semantic priors from a query-level memory bank to strengthen the current-frame representations. We then apply Temporal Consistency Denoising (TCD), a local temporal-window mechanism that denoises unreliable queries, suppresses unstable predictions, and promotes semantic coherence across frames. Finally, the denoised queries are fed into the mask prediction head to produce the segmentation masks.

\begin{figure}
    \centering
    \includegraphics[width=1.0\linewidth]{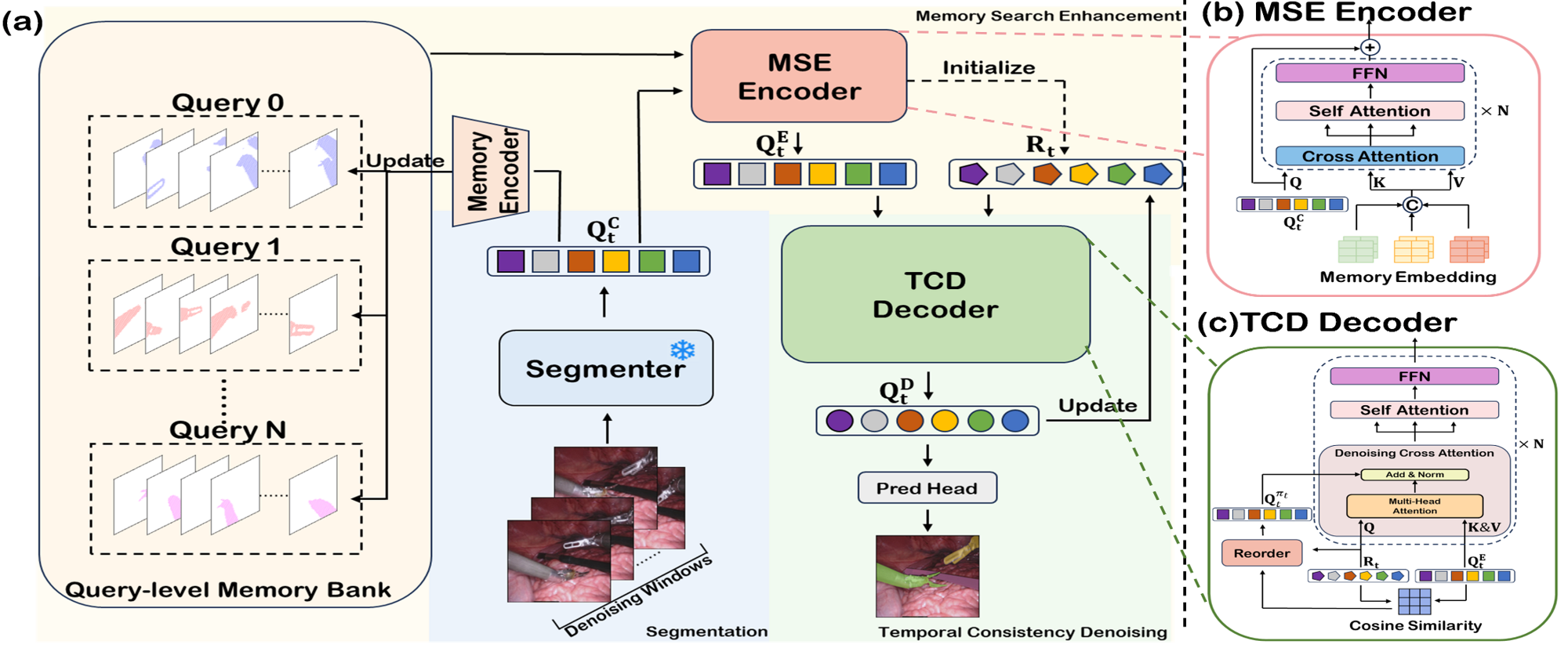}
    \caption{The overview of TEDi. Panel (a) illustrates the overall pipeline of TEDi, while (b) and (c) depict the architectures of the MSE Encoder and the TCD Decoder.
}
    \label{fig:placeholder}
\end{figure}

\subsection{Preliminaries}

\subsubsection{Baseline Segmentation Model.}

We adopt Mask2Former~\cite{cheng2022masked} as the baseline, a representative query-based segmentation framework that formulates segmentation as predicting categories and masks for a fixed set of learnable object queries. Given an input image $I\in\mathbb{R}^{H\times W}$, the backbone and pixel decoder produce a feature map $F$, which interacts with $N$ object queries $\mathbf{Q}=\{\mathbf{q}_i\}_{i=1}^{N},
\mathbf{q}_i\in\mathbb{R}^{d}$ through a transformer decoder. Each query outputs a mask prediction $\mathbf{m}_i\in[0,1]^{H\times W}$ and a classification score $\mathbf{c}_i\in\mathbb{R}^{C}$, forming a set of segmentation candidates.

\subsection{Proposed Methodology}

\subsubsection{Memory Search Enhancement Encoder.}

Memory mechanisms are widely adopted in video segmentation to enhance current predictions using representations from previous frames~\cite{ravi2024sam,liu2024surgical}. However, storing dense pixel-level features often introduces redundancy and weakens retrieval effectiveness~\cite{cheng2022xmem,zhou2024rmem}. We instead propose a \textbf{query-level memory bank} with a \textbf{memory search enhancement encoder}, storing compact instance-aware queries. By encoding predicted masks and injecting mask-aware representations into the queries, the memory remains both efficient and spatially informative.

Firstly, we fine-tune the segmenter and then freeze its parameters
to produces a set of query-level predictions pairs $\mathcal{H}_t=$ 
\(\{\mathbf{q}_{t,i},\mathbf{c}_{t,i},\mathbf{m}_{t,i}\}_{i=1}^{N}\) for frame $t$.

Secondly, we encode the mask \(\mathbf{m}_{t,i}\) into the query embedding space:
\begin{equation}
\mathbf{z}_{t,i}=\phi_m(\mathbf{m}_{t,i}) \in\mathbb{R}^{d} ,
\end{equation}
and fuse it with the original query $\mathbf{q}_{t,i}$ using a learnable gating function to obtain the query-level memory:
\begin{equation}
\mathbf{e}_{t,i}
=
\mathbf{g}_{t,i}\odot\mathbf{q}_{t,i}
+
(1-\mathbf{g}_{t,i})\odot\mathbf{z}_{t,i},
\quad
\mathbf{g}_{t,i}=\sigma\!\left(\phi_g(\mathbf{q}_{t,i},\mathbf{z}_{t,i})\right) \in\mathbb{R}^{d} ,
\end{equation}
where $\phi_m(\cdot)$ and $\phi_g(\cdot)$ denote learnable projection functions, 
$\sigma(\cdot)$ is the sigmoid function, and $\odot$ denotes element-wise multiplication.
Hence, the memory of frame $t$ is defined as $\mathbf{E}_t=\{\mathbf{e}_{t,i}\}_{i=1}^{N} \in \mathbb{R}^{N\times d}$.

Thirdly, a fixed-capacity memory bank $\mathcal{M}_t$
containing query embeddings from the recent $K$ frames is maintained by updating online in a FIFO manner.
\begin{equation}
\mathcal{M}_t
=
\{\mathbf{E}_\tau\}_{\tau\in\mathcal{T}_t}, 
\quad
\mathcal{T}_t
=
\{\tau \mid t-K\le \tau \le t-1\}.
\end{equation}

Finally, given the current-frame queries
\(
\mathbf{Q}_t^{C}=\{\mathbf{q}_{t,i}\}_{i=1}^{N}\in\mathbb{R}^{N\times d}
\) 
and the concatenated stored memories along the query dimension
\(
\mathbf{M}_t
=
\mathrm{Stack}\big(\mathcal{M}_t\big)
\in\mathbb{R}^{KN\times d}
\), we obtain enhanced queries by retrieving cross-frame semantic priors via stacked memory attention blocks:
\begin{equation}
{\mathbf{Q}}_t^{E}
=
\mathrm{MemAttn}(\mathbf{Q}_t^{C},\mathbf{M}_t).
\end{equation}

The enhanced queries ${\mathbf{Q}}_t^{E}=\{{\mathbf{q}}_{t,j}^{E}\}_{j=1}^{N}\in\mathbb{R}^{N\times d}$ are then used for subsequent temporal denoising and prediction.

\subsubsection{Temporal Consistency Denoising Decoder.}
The baseline segmenter is trained in a frame-wise manner and therefore lacks explicit temporal constraints. When applied to videos, its query predictions may become temporally inconsistent or overly confident yet incorrect. Such instability can propagate across frames and impair segmentation reliability~\cite{wang2025LACOSTE}. To address this limitation, we introduce a \textbf{temporal consistency denoising decoder} that refines query embeddings through identity-aligned temporal correction, enforcing coherence across frames and suppressing per-frame noisy query representations.

For frame $t$, we maintain a Temporal Consistency Reference $\mathbf{R}_t = \{\mathbf{r}_{t,i}\}_{i=1}^{N}\in\mathbb{R}^{N\times d}$, which is defined as the denoised queries from the previous frame and serves as a temporally stable semantic anchor:
\begin{equation}
\mathbf{R}_t =
\begin{cases}
{\mathbf{Q}}_{t}^{E}, & t=0,\\
{\mathbf{Q}}_{t-1}^{D}, & t>0.
\end{cases}
\end{equation}

Then, we evaluate the similarity between the Reference and the enhanced queries, and obtain a one-to-one assignment $\pi_t$ via the Hungarian algorithm:
\begin{equation}
\pi_t=\arg\max_{\pi}\sum_{i=1}^{N}\mathbf{S}_{i,\pi(i)},
\quad
\mathbf{S}_{i,j}
=
\frac{\langle \mathbf{r}_{t,i},{\mathbf{q}}_{t,j}^{E}\rangle}
{\|\mathbf{r}_{t,i}\|_2\,\|\mathbf{q}^{E}_{t,j}\|_2}.
\end{equation}

Hence, the enhanced queries can be subsequently reordered to align with $\mathbf{R}_t$:
\begin{equation}
{\mathbf{Q}}_t^{\pi_t}
=
\mathrm{Reorder}({\mathbf{Q}}_t^{E},\pi_t).
\end{equation}

Finally, we refine the queries with $L$ cascaded denoising blocks using temporal consistency cues, where the core denoising cross-attention is defined as:

\begin{equation}
\widetilde{\mathbf{Q}}_t^{\ell}
=
\widetilde{\mathbf{Q}}_t^{\ell-1}
+
\mathrm{CA}(\mathbf{R}_t, \mathbf{Q}_t^{E}, \mathbf{Q}_t^{E}),
\qquad
\widetilde{\mathbf{Q}}_t^{0}=\mathbf{Q}_t^{\pi_t},
\end{equation}
here, the residual term is applied on $\widetilde{\mathbf{Q}}_{t}^{\ell-1}$, ensuring that the decoder preserves the current-frame representation and only injects temporally consistent information. Within each block, we further apply self-attention and a FFN Network.

After $L$ iterations, the denoised queries become the output of the decoder
$\mathbf{Q}_t^{D} =\widetilde{\mathbf{Q}}_t^{L}$, as well as the reference $\mathbf{R}_{t+1}$ for the next frame.
To prevent long-term error accumulation, the temporal reference is reset every $T$ frames. 

\subsubsection{Optimization.} 
We train TEDi using a classification loss $\mathcal{L}_{Cls}$ and a mask prediction loss $\mathcal{L}_{Mask}$. Specifically, we adopt focal loss~\cite{lin2017focal} for $\mathcal{L}_{Cls}$, and combine cross-entropy (CE) loss and Tversky loss~\cite{salehi2017tversky} for mask supervision. The overall loss is defined as:
\begin{equation}
\mathcal{L}_{total}
=
{\lambda_1}\mathcal{L}_{Cls}+{\lambda_2}\mathcal{L}_{ce}+{\lambda_3}\mathcal{L}_{Tversky},
\end{equation}
where $\lambda_1$, $\lambda_2$, and $\lambda_3$ denote the loss weights.
In practice, we set $\lambda_1 = 2$, $\lambda_2 = 5$, and $\lambda_3 = 5$.

\section{Experiments}

All experiments were conducted using publicly available datasets. We evaluate TEDi on two public benchmarks for surgical instrument segmentation,  EndoVis 2017(EV17)~\cite{allan20192017} and EndoVis 2018(EV18)~\cite{allan20202018}. To ensure fair comparisons with prior work, we strictly follow the commonly adopted dataset splits and label settings used in previous studies. For EV17, we follow \cite{shvets2018automatic} and report results under the 4-fold cross-validation protocol. For EV18, We adopt the validation splits and the instrument category annotations provided in~\cite{gonzalez2020isinet}
\begin{table}[ht]
\centering
\caption{Performance comparison on EV17/18 datasets. M2F refers to the baseline Mask2Former model, while $N$ denotes the number of learnable object queries.
}
\label{tab:endo_vis_17_18}
\newcolumntype{C}[1]{>{\centering\arraybackslash}p{#1}}
\renewcommand{\arraystretch}{0.7}  
\setlength{\tabcolsep}{3pt}         
\begin{tabular*}{\textwidth}{
@{\extracolsep{\fill}}
p{7.2em}  
C{2.2em}  
C{2.2em}  
C{2.2em}  
C{2.2em}  
C{2.2em}  
C{2.3em}  
C{2.6em}  
C{2.9em}  
C{2.3em}  
C{2.2em}  
@{}
}
\toprule
\multirow{2}{*}{Model}
& \multirow{2}{*}{\makecell{Ch\_\\IoU}}
& \multirow{2}{*}{\makecell{ISI\_\\IoU}}
& \multirow{2}{*}{\makecell{mc\_\\IoU}}
& \multicolumn{7}{c}{Instrument Categories IoU} \\
\cmidrule{5-11}
&  &  &  & BF & PF & LND & VS/SI & GR/CA & MCS & UP \\
\midrule
\multicolumn{11}{c}{Dataset EV17} \\
\midrule

TernausNet\cite{shvets2018automatic} & 35.27 & 39.87 & 14.19 & 44.20 & 4.67 & 0.00 & 0.00 & 0.00 & 50.44 & 0.00 \\
Dual-MF\cite{zhao2020learning} & 45.80 & -     & 26.40 & 34.40 & 21.50 & 64.30 & 24.10 & 0.80 & 17.90 & 21.80 \\
ISINet\cite{gonzalez2020isinet} & 55.62 & 52.20 & 28.96 & 38.70 & 38.50 & 50.09 & 27.43 & 2.01 & 28.72 & 12.56 \\
S3Net\cite{baby2023forks} & 72.54 & 71.99 & 46.55 & 75.08 & 54.32 & 61.84 & 35.50 & 27.47 & 43.23 & 28.38 \\
MATIS(Full)\cite{ayobi2023MATIS} & 71.36 & 66.28 & 41.09 & 68.37 & 53.26 & 53.55 & 31.89 & 27.34 & 21.34 & 26.53 \\
LACOSTE(S)\cite{wang2025LACOSTE} & 76.32 & 72.37 & 48.22 & 73.24 & 52.04 & 60.41 & 38.73 & 0.00 & 54.53 & 67.88 \\
QPD\cite{dhanakshirur2023learnable} & 77.80 & \textBF{79.58} & 49.92 & 70.61 & 45.84 & \textBF{80.01} & \textBF{63.41} & \textBF{33.64} & \textBF{66.57} & 35.28 \\
\cmidrule{1-11}
M2F($N=100$)  & 75.12 & 71.68 & 44.48 & 60.42 & \textBF{62.97} & 60.88 & 36.29 & 3.14 & 30.22 & 44.48 \\
M2F($N=10$)  & 77.79 & 73.05 & 49.63 & \textBF{76.33} & 60.06 & 65.06 & 35.69 & 3.81 & 42.75 & 63.68 \\
TEDi        & \textBF{80.71} & 78.30 & \textBF{52.11} & 75.55 & 58.55 & 65.78 & 39.32 & 15.76 & 40.64 & \textBF{69.13} \\
\midrule

\multicolumn{11}{c}{ Dataset EV18} \\
\midrule

TernausNet\cite{shvets2018automatic} & 35.27 & 12.67 & 10.17 & 13.45 & 12.39 & 20.51 & 5.97 & 1.08 & 1.00 & 16.76 \\
Dual-MF\cite{zhao2020learning}     & 70.40 & -     & 35.09 & 74.10 & 6.80 & 46.00 & 30.10 & 7.60 & 80.90 & 0.10 \\
ISINet\cite{gonzalez2020isinet}      & 73.03 & 70.97 & 38.73 & 73.83 & 40.35 & 30.98 & 37.68 & 0.00 & 88.16 & 0.16 \\
S3Net\cite{baby2023forks}        & 75.81 & 74.02 & 42.58 & 77.22 & 50.87 & 19.83 & 50.59 & 0.00 & 92.12 & 7.44 \\
MATIS(Full)\cite{ayobi2023MATIS} & 84.26 & 79.12 & 54.04 & 83.52 & 41.90 & 66.18 & \textBF{70.57} & 0.00 & 92.96 & \textBF{23.12} \\
LACOSTE(S)\cite{wang2025LACOSTE}  & 85.20 & 82.41 & 55.92 & 85.21 & \textBF{70.75} & 68.02 & 62.64 & 12.81 & 91.98 & 0.00 \\
QPD\cite{dhanakshirur2023learnable} & 77.77 & 78.43 & 43.84 & 82.80 & 60.94 & 19.96 & 49.70 & 0.00 & 93.93 & 0.00 \\
\cmidrule{1-11}
M2F($N=100$) & 82.93 & 80.54 & 49.77 & 85.23 & 69.42 & 45.37 & 56.35 & 0.00 & 91.99 & 0.00 \\
M2F($N=10$)   & 84.18 & 81.26 & 48.91 & 86.90 & 43.47 & 60.46 & 54.00 & 4.27 & 93.29 & 0.00 \\
TEDi        & \textBF{86.69} & \textBF{84.82} & \textBF{57.41} & \textBF{87.45} & 56.22 & \textBF{69.91} & 69.96 & \textBF{14.82} & \textBF{93.94} & 9.59 \\

\bottomrule
\end{tabular*}
\end{table}

We report performance using three commonly used metrics in this field: challenge IoU (Ch\_IoU), ISINet IoU(ISI\_IoU), and mean class IoU (mc\_IoU).

\subsection{Implementation Details}

We use Swin-Small~\cite{liu2022swin} as the backbone and set the number of object queries to $N=10$ to reduce redundancy. The memory bank capacity  and  denoising window length are set to $K=3$ and $T=3$, respectively. Memory bank initialized from 3 preceding frames. Training is performed with Adam at a learning rate of $6\times10^{-5}$. Inference is restricted to the last frame of each window to maintain strict causality. All experiments are conducted on a single NVIDIA A100 GPU.


\subsection{Comparison with State-of-the-Art Methods}
\begin{figure}
    \centering
    \includegraphics[width=1.0\linewidth]{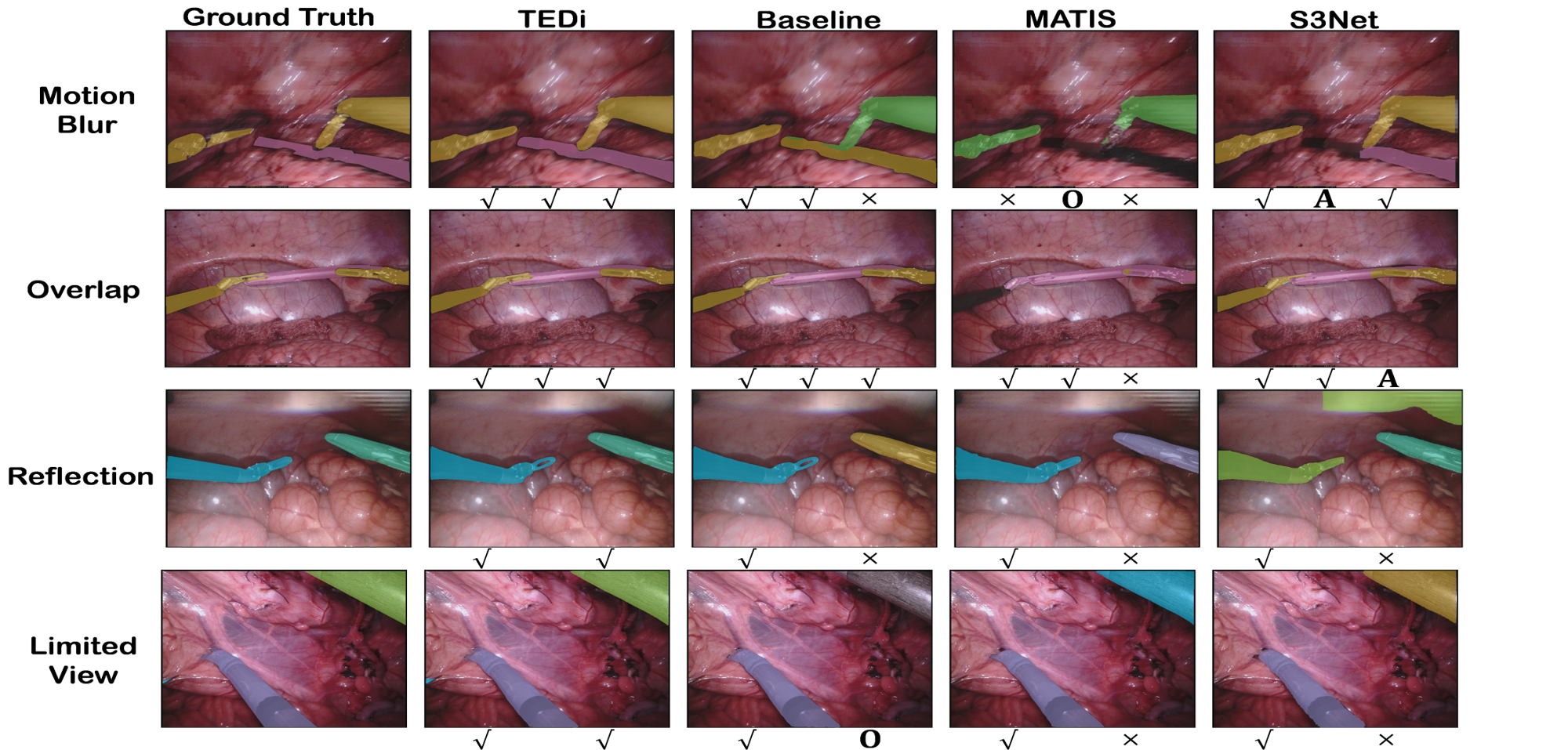}
    \caption{Qualitative comparison of TEDi with other SOTA methods. $\checkmark$ indicates that the instrument is classified and segmented correctly. $\mathbf{O}$ indicates missed instance, $\times$ shows the misclassified instance and $\mathbf{A}$ indicates ambiguous instance.}
    \label{fig:qual}
\end{figure}

We compare TEDi with representative approaches, including single-frame models (QPD~\cite{dhanakshirur2023learnable}, S3Net~\cite{baby2023forks}) and temporal models (MATIS~\cite{ayobi2023MATIS}, LACOSTE~\cite{wang2025LACOSTE}).
As shown in Table~\ref{tab:endo_vis_17_18}, TEDi consistently outperforms prior methods on EV18 across all three metrics. Compared with QPD, TEDi improves Ch\_IoU, ISI\_IoU, and mc\_IoU by 8.92, 6.39, and 13.57 points, respectively. Notably, for challenging classes such as Clip Applier and Large Needle Driver, TEDi achieves substantially higher class IoU, demonstrating the effectiveness of cross-frame memory retrieval and temporal-consistency denoising in mitigating category confusion.
Compared with temporal approaches such as LACOSTE and MATIS, TEDi further achieves clear gains in Ch\_IoU and ISI\_IoU, highlighting the benefit of memory interaction and consistency-guided refinement in the query embedding space.
On EV17, TEDi attains 80.71\% Ch\_IoU and 78.30\% ISI\_IoU, surpassing existing state-of-the-art methods on the main metrics. Improvements over LACOSTE and MATIS remain consistent across evaluation protocols, confirming the robustness of the proposed framework.

Fig.~\ref{fig:qual} presents qualitative comparisons. In challenging scenarios with blurred boundaries, partial visibility, or temporal inconsistencies, existing methods often exhibit class flickering or semantic drift, whereas TEDi produces more stable predictions and improved category discrimination for confusable instruments.


\subsection{Query Visualization Analysis}
We further analyze the baseline’s query predictions during inference and identify two major sources of category errors: (i) high-confidence but incorrectly classified queries that introduce noisy semantics, and (ii) correctly classified yet low-confidence queries whose scores are suppressed due to motion blur or occlusion. The latter are prone to being overlooked during decoding, reducing category separability.

\begin{figure}
    \centering
    \includegraphics[width=1.0\linewidth]{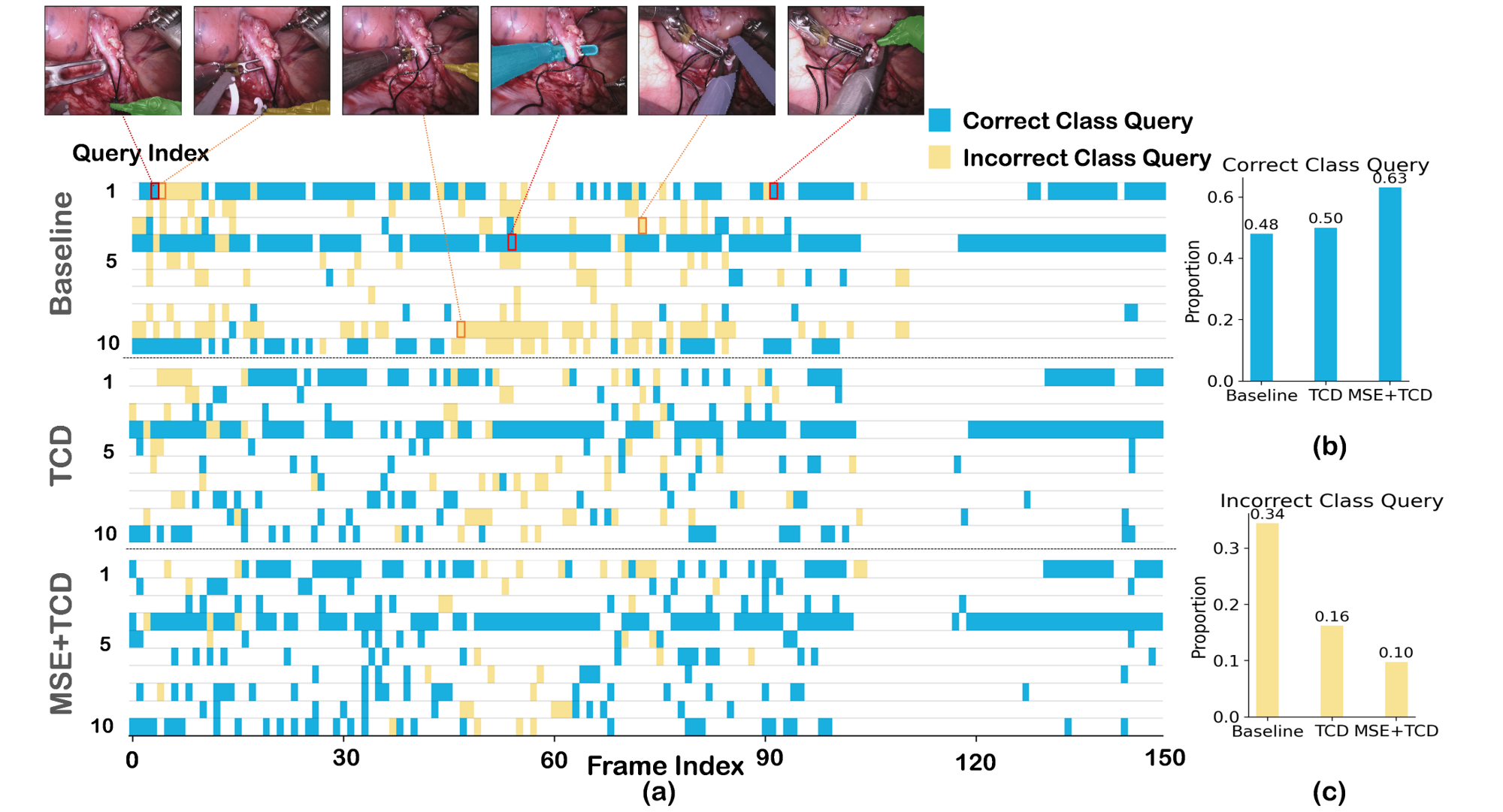}
    \caption{The visualization analysis of query predictions. Panel (a) visualizes whether high-confidence query category predictions are correctly classified on EV18 seq\_2 (frames 000–149). Panel (b) reports the statistics of queries with correct category predictions. Panel (c) reports the statistics of queries with incorrect category predictions.
}
    \label{fig:query_vis}
\end{figure}

To validate the effectiveness of the proposed modules, we visualize the accuracy of high-confidence queries on EV18 seq\_2, which are selected as those with class confidence score $s_{Cls} > 0.8$ and mean class confidence score $s_{Mask} > 0.9$.
As shown in Fig.~\ref{fig:query_vis}, blue denotes correctly classified queries and yellow denotes mismatches. After incorporating MSE and TCD, the number of high-confidence correct queries increases substantially, while high-confidence incorrect queries are markedly reduced. These results suggest that MSE compensates for ambiguous frames via cross-frame semantic retrieval, and TCD suppresses noisy representations through temporal-consistency regularization.

\subsection{Ablation Studies and Hyper-parameter Analysis}


We conduct ablation studies on EV18 to evaluate the contribution of each component and analyze the impact of the memory length $K$ and denoising window size $T$. The results are reported in Table~\ref{tab:ablation}.

Overall, introducing the query-level memory bank with MSE consistently improves category discrimination, while the TCR-guided denoising decoder further enhances temporal semantic consistency and overall segmentation performance. 
In terms of hyper-parameters, the best performance is achieved with $K=3$ and $T=3$. Overly long memory bank or denoising window tends to introduce more historical noise and causes error accumulation over time, whereas an overly short denoising window is insufficient to fully exploit temporal-consistency cues.

\subsection{Computational Efficiency Analysis}To evaluate computational efficiency, we conducted experiments on a single NVIDIA A100 GPU using inputs at a resolution of 512 × 512. TEDi contains 60.58M parameters and requires 78.92 GFLOPs, compared with 56.91 GFLOPs for the Mask2Former baseline. During temporal inference, TEDi reuses query representations from previously encoded frames, such that only the newly arriving frame requires backbone encoding. As a result, TEDi achieves an inference latency of 57.96 ms per clip, with the memory bank introducing only approximately 5.5 ms of additional overhead and a peak GPU memory footprint of 666.1 MB.

\section{Conclusions}
  
We propose TEDi, a temporally enhanced framework for query-based surgical instrument segmentation that integrates MSE and TCD to leverage cross-frame priors and enforce temporal coherence. Experiments on two benchmarks show consistent state-of-the-art performance, improving category prediction stability while remaining efficient and validating the benefits of query-level memory interaction and temporal correction for surgical scene understanding.

\newcolumntype{C}[1]{>{\centering\arraybackslash}p{#1}}

\begin{table}[htbp]
\centering
\caption{
Ablation analysis on EV18 dataset. For column 2-3, $K$ denotes the length of  the memory bank,
while $T$ denotes the denoising window size.
}
\label{tab:ablation}
\renewcommand{\arraystretch}{0.85}  
\setlength{\tabcolsep}{3pt}         

\begin{tabular*}{\textwidth}{
@{\extracolsep{\fill}}
p{4.5em}  
C{1.25em}  %
C{1.25em}  %
C{2.2em}  
C{2.2em}  
C{2.2em}  
C{2.2em}  
C{2.2em}  
C{2.3em}  
C{2.3em}  
C{2.3em}  
C{2.3em}  
C{2.2em}  
@{}
}

\toprule
\multirow{2}{*}{Model}
& \multicolumn{2}{c}{Params.}
& \multirow{2}{*}{\makecell{Ch\_\\IoU}}
& \multirow{2}{*}{\makecell{ISI\_\\IoU}}
& \multirow{2}{*}{\makecell{mc\_\\IoU}}
& \multicolumn{7}{c}{Instrument Categories IoU} \\

\cmidrule{2-3}\cmidrule{7-13}

& $K$ & $T$ &  &  &  & BF & PF & LND & SI & CA & MCS & UP \\
\midrule

M2F(N=10) & - & - & 84.18 & 81.26 & 48.91
& 86.90 & 43.47 & 60.46 & 54.00 & 4.27 & 93.29 & 0.00 \\

MSE   & 3 & - & 85.65 & 83.26 & 51.67
& 88.40 & 54.77 & 65.87 & 52.83 & 0.00 & 94.11 & 5.72 \\

TCD     & - & 3 & 85.59 & 83.07 & 51.70
& 88.16 & 49.51 & 62.86 & 57.94 & 0.00 & 93.90 & 9.51 \\
\cmidrule{1-13}
MSE+TCD & 1 & 5 & 85.65 & 83.69 & 52.56
& 88.00 & \textBF{61.19} & 69.79 & 47.90 & 0.00 & 94.07 & 6.97 \\

MSE+TCD & 2 & 4 & 85.79 & 83.79 & 52.91
& 88.43 & 56.57 & 66.57 & 55.18 & 0.00 & \textBF{94.32} & 9.30 \\

MSE+TCD & 4 & 2 & 86.47 & 84.28 & 55.23
& \textBF{88.80} & 58.47 & \textBF{70.94} & 54.18 & 2.32 & 94.16 & \textBF{17.72} \\

MSE+TCD & 5 & 1 & 85.44 & 83.50 & 51.60
& 88.47 & 56.41 & 67.38 & 50.39 & 0.00 & 94.25 & 4.30 \\

MSE+TCD & 3 & 3 & \textBF{86.69} & \textBF{84.82} & \textBF{57.41}
& 87.45 & 56.22 & 69.91 & \textBF{69.96} & \textBF{14.82} & 93.94 & 9.59 \\
\bottomrule
\end{tabular*}
\end{table}
\subsubsection{\ackname}Jiahong Yuan and Tao Zhang received support from the 7th People's Hospital of Zhengzhou. Weiming Mi and Haoyin Zhou received support from NIH grants R00EB027177 and R01EB036996.
\subsubsection{\discintname}The authors have no competing interests to declare that are relevant to the content of this article.
\bibliographystyle{splncs04}
\bibliography{ref}
\end{document}